# An Agent-based Classification Model

Feng Gu, Uwe Aickelin, Julie Greensmith
School of Computer Science University of
Nottingham
{fxg,uxa,jqg}@cs.nott.ac.uk

## 1 Introduction

The major function of this model is to access the UCI Wisconsin Breast Can- cer data-set[1] and classify the data items into two categories, which are normal and anomalous. This kind of classifi cation can be referred as anomaly detection, which discriminates anomalous behaviour from normal behaviour in computer systems. One popular solution for anomaly detection is Artifi cial Immune Sys- tems (AIS). AIS are adaptive systems inspired by theoretical immunology and observed immune functions, principles and models which are applied to prob- lem solving. The Dendritic Cell Algorithm (DCA)[2] is an AIS algorithm that is developed specifi cally for anomaly detection. It has been successfully applied to intrusion detection in computer security. It is believed that agent-based mod- elling is an ideal approach for implementing AIS, as intelligent agents could be the perfect representations of immune entities in AIS. This model evaluates the feasibility of re-implementing the DCA in an agent-based simulation environ- ment called AnyLogic, where the immune entities in the DCA are represented by intelligent agents. If this model can be successfully implemented, it makes it possible to implement more complicated and adaptive AIS models in the agent-based simulation environment.

## 2 The DCA

Natural dendritic cells (DCs) are professional antigen presenting cells (APCs) that are capable of fusing and processing multiple signals from separate sources. There are three states of DCs' lifecycle which are immature, semimature and fully mature. The DCs' lifecycle starts from immature state, when immature DCs are ingesting the antigens and sensing the signals around them all the time. Based on the combination of antigens and signals, immature DCs can then diff erentiate to semimature DCs or mature DCs. Immature DCs normally encounter three signals which are PAMPs (Pathogen Associated Molecular Pat- terns), danger signals (released by unregulated cell deaths) and safe signals (released by programmed cell deaths). PAMPs and danger signals make DCs toward fully mature state, however safe signals make DCs toward semimature state. The diff erentiation pathways are shown in Figure 1. The DCA is the algorithm that maps the DCs diff erentiation mechanism into computer systems by using quantitative measuring methods. Similar to the natural immunity, PAMP, danger signal and safe signal are also generated as the input signals. Each set of input signals then are processed by the algorithm, in order to get three output signals which are Csm (Costimulation signal), Semi (Semimature signal) and Mat (Mature signal).

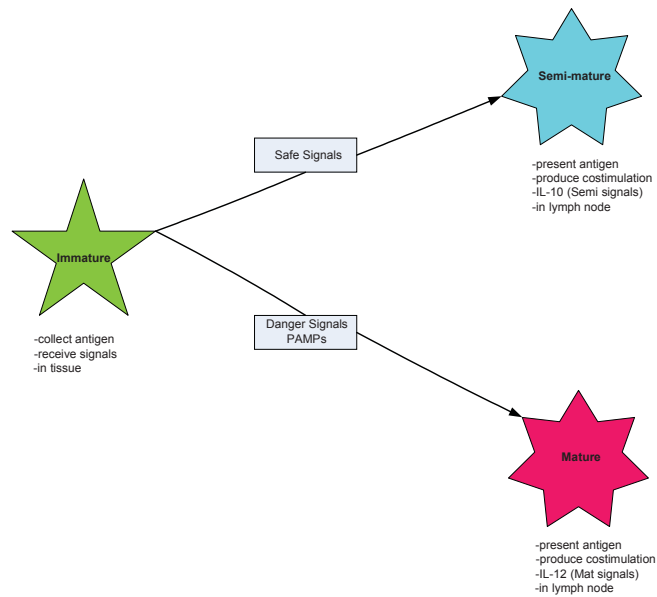

Figure 1: Differentiation pathways of DCs maturation

Due to the fact that natural DCs always sample multiple antigens and multiple sets of signals, the DCA sets an individual assigned migration threshold of each DC to cope with this mechanism. During the signal processing of DCs, the cumulative Csm, Semi and Mat of sampled antigens are calculated and recorded. As soon as cumulative Csm exceeds the migration threshold, the DC gets maturation. But the DCs differentiation direction is determined by the comparison between cumulative Semi and cumulative Mat. If the cumulative Semi is greater than the cumulative Mat, then the DC goes to semimature, otherwise it goes to mature. The semimature DC returns '0' context to the sampled antigens, however the mature DC returns '1' context to the sampled antigens. At the end, each antigen gets a binary string of mature contexts which can be calculated to get the mature context antigen value (MCAV) through the number of context '1' divided by the number of all contexts. It is similar to a voting system, where the antigen is the candidate and the DCs that sampled this antigen are the voters. If the context is '1', it means the DC 'thinks' the antigen is anomalous, whereas if the context is '0', the DC 'thinks' the antigen is normal. Thus the MCAV is actually the probability of that this antigen is being anomalous. An anomalous threshold is also introduced. Those antigens whose MCAV are greater than the anomalous threshold are classified into the anomalous category, while the others are classified into normal category.

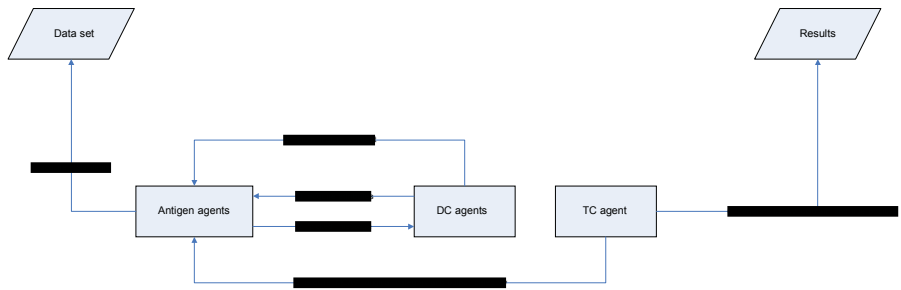

Figure 2: Interactions between agents

## 3 Model Implementation

Three types of agents are created in this model which are Antigen agents, DC agents and TC (T-cell) agent, to represent the entities involved in the DCA. Antigen agents are the data carriers, DC agents are the data processors and TC agent is the statistical analyser of the classification. The TC agent in this model has not been an exact representation of natural T-cells yet, but more biological properties will be added during the future development. AnyLogic is a Java-based simulation software where all the agents are initiated in the Main class. But the individual behaviours and the communications between agents are managed by the agents themselves. An environment is also defined where all the agents can interact with each other, to facilitate the communications between agents. The interactions between agents are shown in Figure 1.

Firstly, one Antigen agent is created per second until all the data items are processed. Each Antigen agent represents one data item of the data set. Antigen agents first extract and record selected attributes from the data items. As one antigen samples multiple times in the DCA, each Antigen agent then randomly selects certain amount of DC agents from the DC population (all the existing DC agents) and sends those DC agents a message 'picked'. When the Antigen agent receives the complete mature contexts from the DC agents, it calculates the MCAV. The MCAV is also compared to the MCAV threshold to classify this Antigen agent.

Secondly, DC agents are the most complex agent class in the model. Certain amount of DC agents are created when the system starts as the initial DC population. Each DC agent has three states which are immature, semimature and mature similarly to natural DCs. The DC agent starts from the immature state where an individual assigned migration threshold is generated. When the DC agent receives a 'picked' message from an Antigen agent, it records the ID of that Antigen agent and copies the stored attributes and executes the signal processing function. The signal processing function generates the input signals (PAMP, danger signal and safe signal) from the acquired attributes. It also calculates the output signals (Csm, Semi and Mat) and the cumulative signals. One DC sampling cycle is completed so far, it will keep repeating as long as the DC agent receives messages from Antigen agents, until the cumulative Csm exceeds the migration threshold. Then the DC agent moves to the decision branch where the diff erentiation pathways are decided. When the DC agent is in either semimature state or mature state, it returns the mature contexts to each Antigen agent it sampled. After then this DC agent is terminated and a new DC agent (also starts from immature state) is created and added into the DC population, thus the DC population is constant during the simulation.

Thirdly, TC agent accesses all the Antigen agents to get the MCAV, so that it can generate an MCAV diagram. The MCAV diagram shows the MCAV distribution as well as the two categories of the data items. The TC agent also compares the output category with the original category of each Antigen agent, to calculate the overall true positive or accuracy of the classifi cation. The overall true positive is considered as the measurement of the system performance.

## 4 Summary and Future Work

Through the simulation experiments this model is able to generate the same results as in[2], thus the basic objective is achieved. This implementation also incorporates some advantages of agent-based simulation approach. Firstly, in agent-based simulation approach, the agents manage their own reactive and proactive behaviours. Each agent can interact with other agents in the environ- ment, to achieve certain objectives[3]. Such a framework is similar to the natural immune system, which makes the implementation more straightforward. Sec- ondly, the simulation software that facilitates the agent communications makes it much easier to implement the interactions among the entities in the DCA. Thus it provides a natural advantage during implementation. Thirdly, the agent statuses and agent interactions can be monitored in real time, and the simula- tion can be paused, speeded up or slowed down according to the requirements. Thus the model is more dynamic and fl exible for experiments.

Only three entities are involved in this model and their behaviours and interactions are quite simple. In the future, we intend to introduce more agents into the systems and make the agents more intelligent. Currently the most interesting attibutes have to be specifi cally selected as signal sources, but the choice of attributes will be dependent on the data sets. An intelligent tissue agent will be developed to automatically select the signal sources. This agent also has the ability to learn based on the judgment of previous classifi cation performances, to make the system perform better in later classifi cations.